# Deep learning reveals the common spectrum underlying multiple brain disorders in youth and elders from brain functional networks

Mianxin Liu[1+,] Jingyang Zhang[1+], Yao Wang[2], Yan Zhou[2], Fang Xie[3], Qihao Guo[4], Feng Shi[5], Han Zhang[1], Qian Wang[1] and Dinggang Shen[1,5,6*]

[1] School of Biomedical Engineering, ShanghaiTech University, Shanghai 201210, China

[2] Department of Radiology, Renji Hospital, School of Medicine, Shanghai Jiao Tong University, Shanghai 200001, China

[3] PET Center, Huashan Hospital, Fudan University, Shanghai 200040, China

[4] Department of Gerontology, Shanghai Jiao Tong University Affiliated Sixth People's Hospital, Shanghai 200233, China

[5] Department of Research and Development, Shanghai United Imaging Intelligence Co., Ltd., Shanghai 200232, China

[6] Shanghai Clinical Research and Trial Center, Shanghai, 201210, China

+: equal contribution

*Corresponding author:
Dinggang.Shen@gmail.com

## Abstract

Brain disorders in the early and late life of humans potentially share pathological alterations in brain functions. However, the key evidence from neuroimaging data for pathological commonness remains unrevealed. To explore this hypothesis, we build a deep learning model, using multi-site functional magnetic resonance imaging data ($N$=4,410, 6 sites), for classifying 5 different brain disorders from healthy controls, with a set of common features. Our model achieves $62.6\pm1.9\%$ overall classification accuracy on data from the 6 investigated sites and detects a set of commonly affected functional subnetworks at different spatial scales, including default mode, executive control, visual, and limbic networks. In the deep-layer feature representation for individual data, we observe young and aging patients with disorders are continuously distributed, which is in line with the clinical concept of the "spectrum of disorders". The revealed spectrum underlying early- and late-life brain disorders promotes the understanding of disorder comorbidities in the lifespan.

## Introduction

The mental health of children and elders is frequently affected by a wide type of brain disorders (BDs), to which prevention, diagnosis, and treatment remain challenging. With the development of research on BDs, the concept of "spectrum" is utilized to integrate different BDs into a unified knowledge framework. Upon a "spectrum", a group of different disorders can share certain features and their symptoms occur on a continuum [1]. Recently, researchers gradually realize that different BDs in early and late life could locate in their respective spectrums. Autism spectrum disorder (ASD) [2,3] and attention-deficit/hyperactivity disorder (ADHD) [4] are two representative BDs in the early life of humans, respectively affecting social interaction and attention abilities in typical cases. Studies have started to explore a potential common spectrum underlying ADHD and ASD [5,6], as they could have similar symptoms, often co-occur with each other [7], and share certain genetic architectures [8]. Meanwhile, mild cognitive impairment (MCI) [9] and dementia often occur in elders, due to Alzheimer's disease (AD) [10], vascular diseases [11,12], and other etiology. Akin to early-life developmental disorders, cognitive impairments in elders cover a broad range of heterogeneous behavior disabilities and are also associated temporally; for instance, vascular cognitive impairments (VCI) often promote the development of AD [11,13]. Thus studies on the commonness among these late-life BDs have emerged, with a hypothesis on the existence of another spectrum underlying late-life BDs [14,15].

Although conventional views regard early development and aging as two dichotomized processes in human life, there is a debate in the past 20 years on a potentially common neurological process shared by them and the associated BDs [16,17]. First, early- and late-life BDs can show similar cognitive-behavioral symptoms. ASD and AD can both manifest memory deficits, language impairment, visuospatial ability decline, and executive function alteration [18]. Second, early- and late-life BDs exhibit strong temporal connections, even though the lapse may span several decades. Patients with ASD can develop into dementia at 2.6 times more likely when compared to the general population [19]. Third, early- and late-life BDs can have common genetic factors involved in their progression. A cohort study in Sweden suggested that ADHD and AD are associated across generations, implying common genetic risks shared by them [20]. The common transcriptomic alteration has also been reported in both early- and late-life BDs [21]. Finally, large-scale neuroimaging studies suggested that ASD, ADHD, AD, and other BDs have the same deficits in the neural functional subsystem in the brain functional networks (BFNs). The "triple networks" are the typical detected common subsystems [22], including the default mode network, executive control network, and saliency network [23–27].

Enlighted by this evidence, when covering ASD, ADHD by an "early-life BD spectrum" and different MCIs and dementias in elders by a "late-life BD spectrum", we ask whether the two spectrums can be integrated so that an approximate "lifespan BD spectrum" spanning 60-70% of the time in the human life can be established? Building such a lifespan BD spectrum could potentially offer a common target for treatment or interference of different BDs and a unified viewpoint to understand the basis of the comorbidities of early- and late-life BDs. However, few studies provide strong neuroimaging evidence for the shared neurological basis and address this issue, while the advance in deep learning (DL) technology provides some promise. Advanced DL models have been applied to analyze different BDs with high sensitivity in feature extractions [28]. In addition, the DL model shows a high capacity to represent large-scale data from different sources within the same model architecture [29]. When hypothesizing a set of common neural features among early- and late-life BDs exist, an advanced DL model is promising to automatically identify the common features by learning from a large amount of neuroimaging data on different BDs, as different assessments of the common information. And the data representation extracted from the model will naturally inform about the feasibility of a lifespan BD spectrum.

In this work, we aim to implement a DL method with promises to investigate the common pathological factor among early- and late-life BDs in the BFN and explore the potential existence of the lifespan BD spectrum. We build a deep learning model based on multiscale BFNs from 4410 functional image data, including 2512 data from healthy controls (HCs), and 1898 data from patients suffering from ASD, ADHD, MCI, AD, or VCI. Specifically-designed bi-classification and transfer learning experiments are performed to demonstrate the existence of common features. Based on deep layer features of the model, we further investigate the data representation space of multiple BDs for exploration on an integrated lifespan BD spectrum.

**A multiscale-BFN-based DL model learns to classify multiple BDs from HCs**

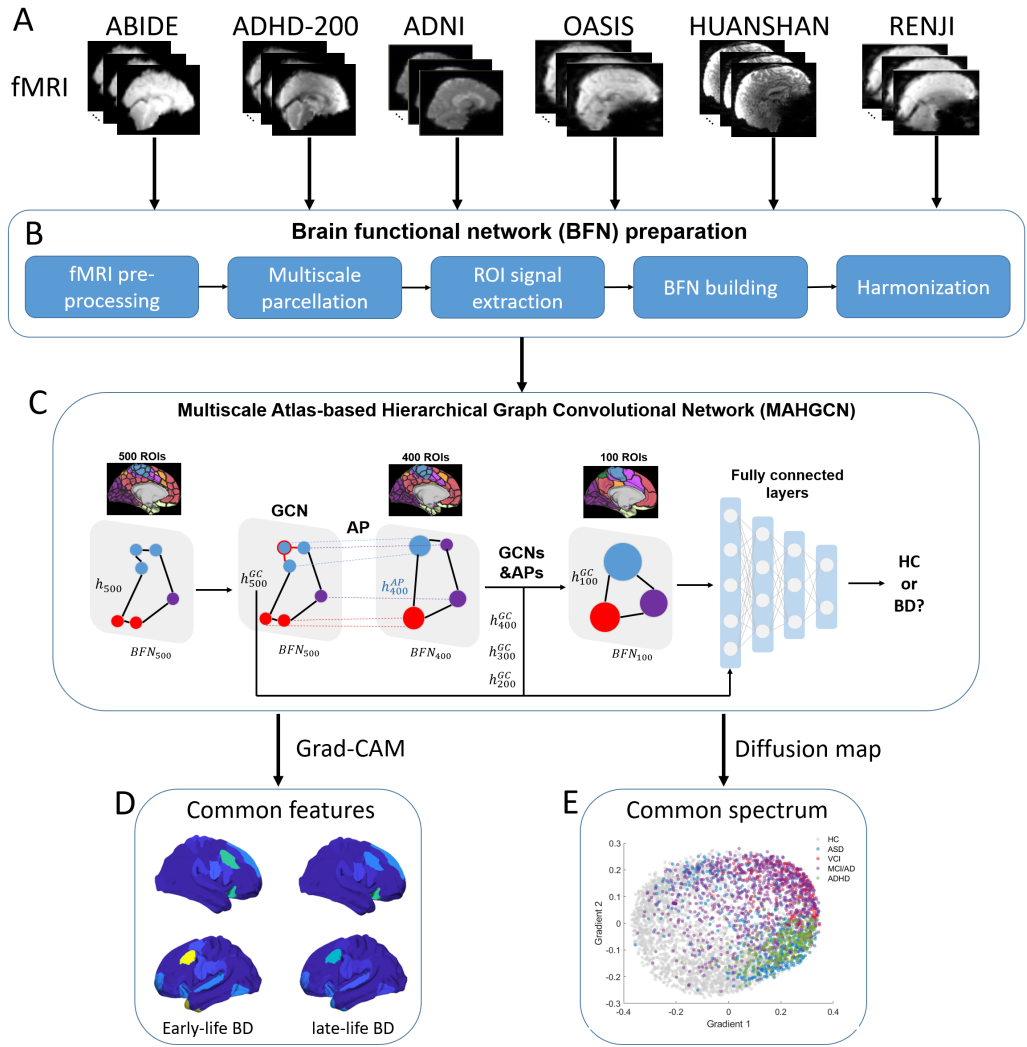

**Figure 1. The workflow of the main analysis pipeline.** (A). The fMRI data from 6 sites. ABIDE and ADHD-200 datasets are respectively for ASD and ADHD studies. ADNI, OASIS, and HUASHAN datasets are for MCI and AD studies. And RENJI dataset is for the VCI study. (B). The BFN preparation steps. The construction of multiscale BFNs is based on the multiscale atlases shown in Fig. S1. (C). The neural network structure of the multiscale atlas-based hierarchical graph convolution network (MAHGCN), which hierarchically extracts and integrates features from multiscale BFNs using stacked graph convolutional networks and atlas-based pooling to make the diagnostic decision. The "BD" group includes ASD, ADHD, MCI, AD, and VCI, and the model performs bi-classification tasks. (D). After the model building, the "Grad-CAM" method is used to explore the common features among BDs encoded inside the MAHGCN model. (E). The "Diffusion map" method is performed to investigate the deep-layer representation of the MAHGCN model for potentially common spectrum

under different BDs.

We applied our previously proposed method, namely multiscale atlas-based hierarchical graph convolution network (MAHGCN) [30], to perform a bi-classification between HCs and various BDs and also conduct a transfer learning experiment on functional neuroimaging data from six sites (Fig. 1A, $N$=4,410). Upon the success of these experiments, we expect to identify common features shared by the classification tasks and provide a unified framework to study the pathology of these BDs.

The MAHGCN analyzing pipeline is shown in Fig. 1C. Briefly, after building the BFNs at different spatial scales based on predefined multiscale atlases and individual fMRI data, the MAHGCN extracts disease-related features from multiscale BFNs, based on stacked GCNs and atlas-guided pooling (AP) operations. Specifically, we implement multiscale atlases from Schaefer *et al.* [31], where brains are parcellated into coarse- and fine-scale regions of interest (ROIs), but a similar correspondence to the seven large-scale resting-state functional networks (RSN) [32] is preserved (Fig. S1). The RSNs include visual network (VIS), somatomotor network (SM), dorsal attention network (DAN), salience network (SAL), limbic network (LIM), executive control network (ECN), and default mode network (DMN). Therefore, the spatial relationships among ROIs in these multiscale atlases can thus be regarded as a biologically meaningful brain hierarchy. Using this prior of the hierarchical relationship between neighboring-scale atlases, the AP is designed to guide nodal feature integration between GCNs. Furthermore, the extracted features from each scale will join the individualized diagnosis decision via skip connections, feature concatenation, and the process of multiple fully-connected layers. In our previous works, we demonstrated the capability of this method in optimally classifying AD, MCI, ASD, and VCI from HC, respectively [30,33,34]. Clues have been achieved by MAHGCN on shared BFN features among MCI, AD, and VCI [30,33,34]. Therefore, MAHGCN can be a promising choice for this work to effectively classify BD subjects from HC and explore the common BFN features among multiple BDs.

Bi-classification experiment

Table 1 and Fig. 2A report the site-averaged classification results (see Method) of the single-scale-based GCNs and MAHGCN during a ten-fold cross-validation. The site-averaged metrics define the average prediction performance on each site and the corresponding task. In general, the single-scale-based GCN achieves 58%-60% accuracy. In contrast, our multi-scale-based MAHGCN obtains an accuracy of 62.6±3.4%, a sensitivity of 61.0±8.6%, a specificity of 66.8±3.5%, and an AUC of 63.9±4.3%. The performance from MAHGCN is significantly higher than the single-scale-based GCN (Table 1).

**Table 1. Site-averaged performances of the bi-classification experiment by different methods.** Bold indicates the highest performance. The symbol * indicates a significantly higher performance of MAHGCN than the single-scale-based GCN at a significance level of $p<0.05$. **: $p<0.01$ and ***: $p<0.001$ after FDR correction. The one-sided Wilcoxon signed-rank test is used to access the significance.

| Method | ACC (%) | $p$ | SEN (%) | $p$ | SPE (%) | $p$ | AUC (%) | $p$ |
|---|---|---|---|---|---|---|---|---|
| 100 ROIs | 58.2±2.3** | 0.0078 | 56.9±3.2 | 0.0967 | 60.4±5.3 | 0.0560 | 58.6±2.1** | 0.0039 |
| 200 ROIs | 59.8±2.0** | 0.0039 | 56.2±3.5 | 0.0967 | 63.5±5.0 | 0.0703 | 59.8±2.0** | 0.0020 |
| 300 ROIs | 59.6±2.7* | 0.0137 | 55.3±3.2 | 0.1289 | 64.1±4.7 | 0.1162 | 59.7±2.2** | 0.0039 |
| 400 ROIs | 58.7±2.3** | 0.0039 | 57.6±5.2 | 0.1377 | 61.0±4.7* | 0.0247 | 59.3±2.3** | 0.0020 |
| 500 ROIs | 58.8±2.8* | 0.0137 | 58.3±3.3 | 0.1611 | 59.6±5.4** | 0.0091 | 58.9±2.4** | 0.0039 |
| MAHGCN | **62.6**±1.9 | N/A | **61.0**±6.8 | N/A | **66.8**±5.3 | N/A | **63.9**±1.8 | N/A |

**Table 2. Site-specific performances of MAHGCN on the bi-classification experiment.** * indicates the performance metrics are significantly higher than the chance level at a significance level of $p<0.05$. **: $p<0.01$ and ***: $p<0.001$ after FDR correction. The one-sided Whitney-Mann's U test is used to access the significance.

| Site | ACC | $p$ | SEN | $p$ | SPE | $p$ | AUC | $p$ |
|---|---|---|---|---|---|---|---|---|
| ABIDE | 57.4±2.5*** | 8.36e-06 | 48.3±14.1* | 0.0228 | 67.4±12.5* | 0.025 | 57.8±2.4*** | 7.01e-06 |
| RENJI | 67.4±9.7*** | 6.59e-06 | 62.9±16.4*** | 7.43e-04 | 73.4±8.9*** | 4.04e-05 | 68.2±9.5*** | 7.02e-06 |
| HUASHAN | 62.6±6.7*** | 7.12e-04 | 61.9±16.1** | 0.0054 | 63.9±8.3* | 0.0135 | 68.9±7.8*** | 4.40e-04 |
| ADNI | 60.7±3.7*** | 1.32e-06 | 57.2±5.6*** | 4.33e-05 | 65.3±5.6*** | 7.96e-06 | 61.2±3.4*** | 1.63e-06 |
| OASIS | 65.0±6.2** | 0.0011 | 74.9±12.6*** | 3.52e-06 | 63.9±7.1* | 0.0394 | 69.4±6.1*** | 2.93e-06 |
| ADHD-200 | 58.7±5.1*** | 2.53e-04 | 58.9±16.0* | 0.0252 | 58.3±12.8 | 0.078 | 58.6±4.8*** | 3.86e-05 |

In addition, Table 2 and Fig. 2B offer details of the site-specific diagnostic performance of MAHGCN. Since the HC-to-BD class ratio can fluctuate and be imbalanced in certain sites, we regard sensitivity and AUC as more informative metrics on predictability. For MCI, AD and VCI, our method obtains AUCs of 68.2±9.5%, 68.9±7.8%, 61.2±3.4%, and 69.4±6.1% for RENJI, HUASHAN, ADNI and OASIS datasets, respectively. And for ASD and ADHD, the model results in lower AUCs than MCI, AD, and VCI, with AUCs of 57.8±2.4% for ABIDE and 58.6±4.8% for ADHD-200. In terms of sensitivity, the model achieves 48.3±14.1%, 62.9±16.4%, 61.9±16.1%, 57.2±5.6% 74.9±12.6%, and 58.9±16.0% for ABIDE, RENJI, HUASHAN, ADNI, OASIS, and ADHD-200. According to permutation tests, the predictabilities in terms of accuracy, sensitivity, and AUC are all significantly higher than the chance level (Table 2). Overall, these results validate a certain level of capability of MAHGCN in diagnosing multiple BDs using a unified neural network architecture, which suggests the existence of common features among multiple BDs in BFNs.

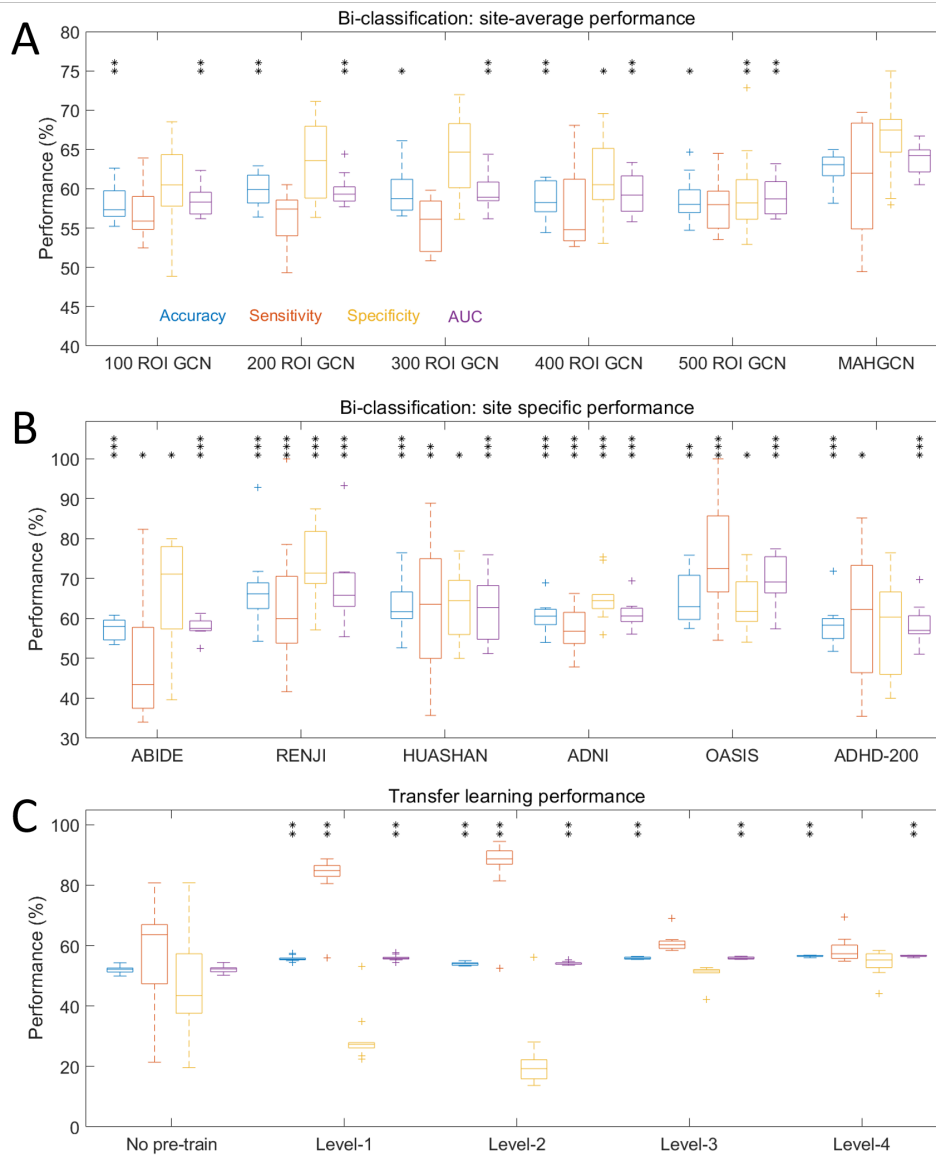

**Figure 2. The detailed distributions of performances in different prediction experiments.** (A). Boxplots for site-averaged performances of the bi-classification experiment by different methods. The symbol * indicates a significantly higher performance of MAHGCN than the single-scale-based GCN at a significance level of $p$<0.05. **: $p$<0.01 and ***: $p$<0.001 after FDR correction. The one-sided Wilcoxon

signed-rank test is used to access the significance. (B) Boxplots for site-specific performances of MAHGCN on the bi-classification experiment. *: the performance metrics are significantly higher than the chance level at a significance level of p<0.05. **: p<0.01 and ***: p<0.001 after FDR correction. The one-sided Whitney-Mann's U test is used to access the significance. (C). Boxplots for the prediction performances in transfer learning experiments under no pre-training and different transfer learning schemes. *: the performance metrics are significantly higher than the baseline performance at a significance level of $p$<0.05. **: $p$<0.01 and ***: $p$<0.001 after FDR correction. The one-sided Wilcoxon signed-rank test is used to access the significance.

Transfer learning experiment

**Table 3. The prediction performances in transfer learning experiments under no pre-training (baseline) and different transfer learning schemes.** The results are from the 20-shot condition. Bold indicates the highest performance. * indicates the performance metrics are significantly higher than the baseline performance at a significance level of $p$<0.05. **: $p$<0.01 and ***: $p$<0.001 after FDR correction. The one-sided Wilcoxon signed-rank test is used to access the significance.

| Scheme | ACC | $p$ | SEN | $p$ | SPE | $p$ | AUC | $p$ |
|---|---|---|---|---|---|---|---|---|
| Baseline | 52.09±1.2 | N/A | 58.51±17.5 | N/A | 45.78±17.6 | N/A | 52.14±1.2 | N/A |
| Level-1 | 55.75±0.9** | 0.0039 | 82.26±9.6** | 0.0020 | 29.66±8.9 | 0.9971 | 55.96±0.9** | 0.0013 |
| Level-2 | 53.97±0.6** | 0.0039 | **85.61±12.2** | 0.0039 | 22.84±12.4 | 0.9990 | 54.22±0.5** | 0.0026 |
| Level-3 | 55.85±0.4** | 0.0039 | 61.02±3.1 | 0.5674 | 50.76±3.1 | 0.2148 | 55.89±0.4** | 0.0020 |
| Level-4 | **56.50±0.3** | 0.0039 | 58.67±4.4 | 0.6875 | **54.36±4.3** | 0.0872 | **56.51±0.3** | 0.0020 |

We further collect evidence for the shared features among BDs using additional transfer learning experiments. A MAHGCN model is pre-trained using all datasets except ABIDE to learn features for MCI, AD, VCI, and ADHD. If the predictive features are shared, the model is expected to possess certain knowledge about ASD and can be fast transferred to identify ASD by fine-tuning the model parameters with a small number of training samples (e.g., $N$=20) from ABIDE dataset. And the pre-trained model should exhibit significantly higher performance than a model without pre-training. Different transfer learning schemes are designed to preserve different levels of learned features during the pre-training (see Methods). A higher-level scheme allows less tuning of the parameters and keeps more learned information. In Table 3 and Fig 2C, the predictability in terms of accuracy and AUC generally increases with the level of preservation of the learned features. And all pre-trained model gives higher accuracies and AUCs than the non-trained model. The Level-4 scheme provides significantly higher accuracy and AUC than the baseline, along with relatively balanced and stable sensitivity and specificity. The level-3 scheme also exhibits increased accuracy, sensitivity, and AUC by roughly 3% when compared to the baseline. In addition, it can be also noted that, in Levels 1 and 2, tuning the parameter with less preservation of the pre-trained information will significantly degrade the specificity and case-imbalanced sensitivity and specificity. Results using 50 and 100 training samples from ABIDE are shown in Tables S2-3, which also indicate that higher levels of preservation of learned information lead to better performances. All observations support diagnostic feature sharing of the MCI, AD, VCI, and ADHD with ASD.

**The common features encoded by DL models indicate shared neurological factors among early- and late-life BDs**

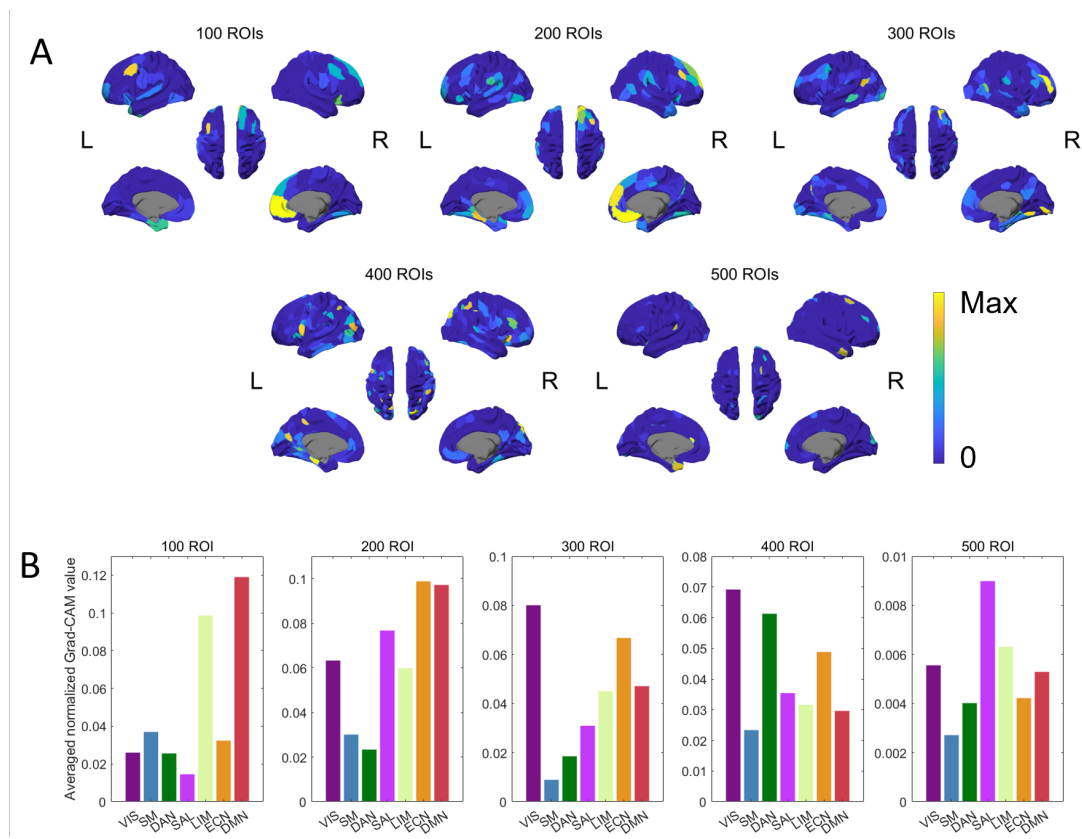

**Figure 3. Shared diagnostic brain regions and RSNs for all involved brain disorders learned by the deep learning model.** (A). The brain maps showing common diagnostic regions identified by high Grad-CAM values; (B) Bar plots for common diagnostic RSNs indicated by RSN-wise averaged Grad-CAM values. The detailed data distributions behind B are offered in Fig. S2A.

We further explore the encoded features in the established models from bi-classification experiments, aiming to capture the commonness of various BDs with different etiology in BFNs (Fig. 1D). We apply the grad-CAM method to the DL model and evaluate the features. A high Grad-CAM score indicates a high contribution to the prediction. Joint consideration of features learned in BD populations by models from cross-validations is achieved by using a series of normalization and weighted averaging. As the Grad-CAM values can vary significantly across different models and different datasets due to the multi-site nature, a normalization in value range is, respectively, applied to the Grad-CAM from different models and different datasets before averaging (see Method).

First, the normalized Grad-CAM values overall data are averaged to estimate the all-BD common features. In Fig. 3A, the detected diagnostic brain regions are not identical using different spatial scales. For 100- and 200-ROI scales, the predictive regions appear in the frontal cortex, while, at 300- and 400-ROI scales, the features are distributed but are largely located in the parietal cortex. Using the language of brain RSNs (Fig. 3B, the RSN-wise averaged Grad-CAM value is used to evaluate the predictability of each RSN), the DMN and LIM are the most predictive features for multiple BDs at 100-ROI scales. At the 200-ROI scale, the model relies more on DMN, ECN, and SAL. For 300- and 400-ROI scales, the model regards VIS, VAN, and ECN as the common diagnostic features, while, in the 500-ROI scale, the SAL is highlighted. Note that though we introduced the skip connections in MAHGCN, the gradient value can still significantly drop in the shallower layers, which influences the inter-scale comparison (Fig. 3B). For instance, the 500-ROI BFN is processed by the shallowest GCN layer and is thus weighted with the least gradient values and scored with the least Grad-CAM values.

In Fig. 4 and Figs. S3-S4, we investigate the features identified by our model for early- and late-life BDs separately, as the common features in Fig. 3 are a mixture of the contributions from different disorders and cannot directly suggest commonness of early- and late-life BDs. In Figs. 4A and 4C, the brain maps suggest that, from 100-ROI to 400 ROI scales, the locations of diagnostic brain regions in early- and late-life BDs are quite consistent, despite certain variations in the amplitudes. There is less consistency in the brain regions on a 500-ROI scale. Furthermore, in Fig. 4B and 4D, it can be observed that the features for early- and late-life BDs at 100-ROI scales are consistent with the all-BD estimation, which regards the DMN and LIM as the most informative RSNs. For the 200-ROI scale, the common RSNs are DMN and ECN, but note that the early-life BDs evaluate more on the SAL and LIM. The feature distributions at 300- and 400-ROI scales are relatively stable and consistent with the all-BD estimation, but VIS and ECN can be regarded as commonness. At 500-ROI scales, estimations from different disorders exhibit large variations and identify crucial RSNs (different from the all-BD estimation).

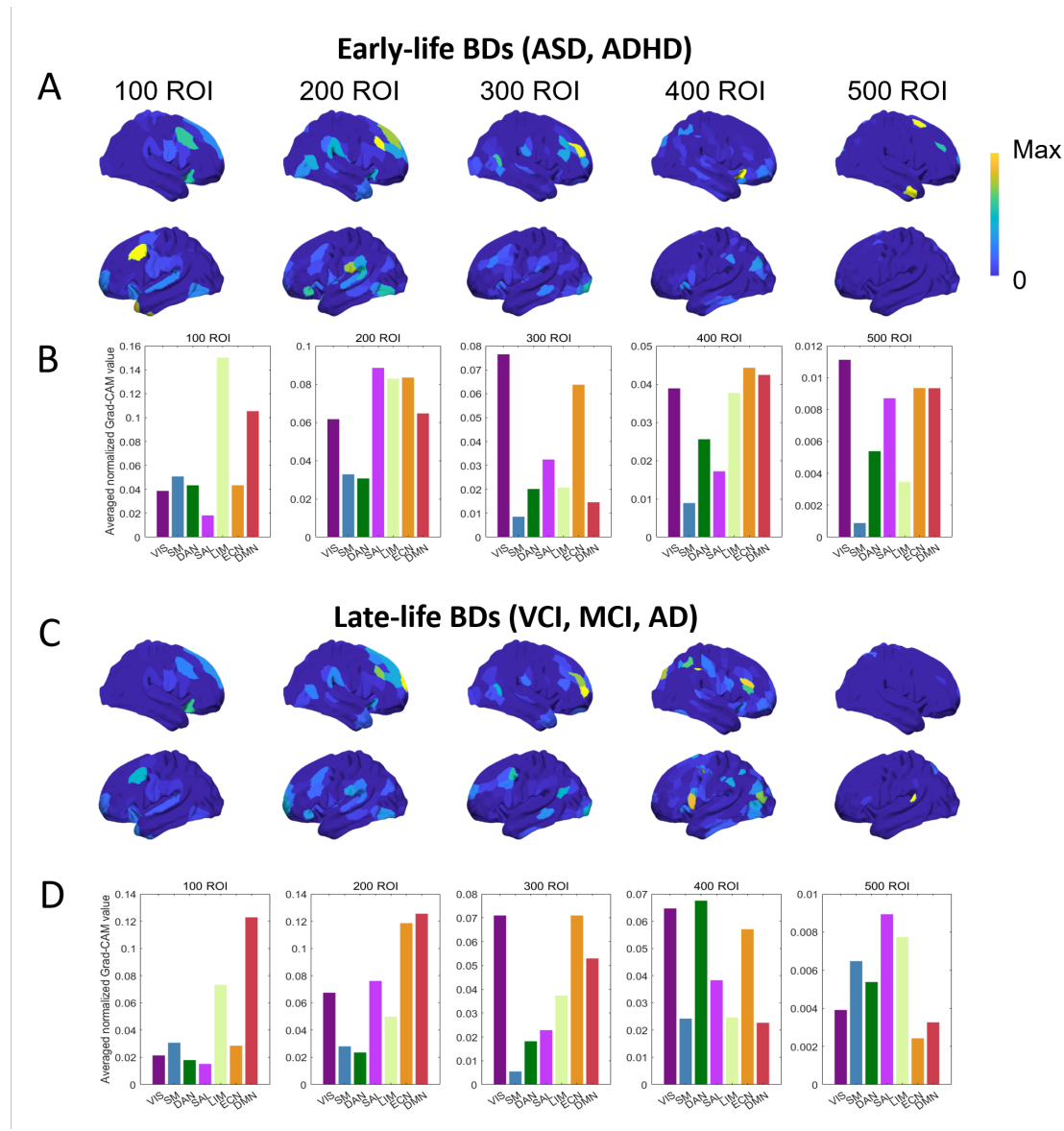

**Figure 4. Diagnostic brain regions and RSNs for early-life and late-life BDs, respectively.** (A-B) The brain maps show common diagnostic regions and bar plots for common diagnostic RSNs for early-life BDs (ASD and ADHD data from ABIDE and ADHD-200 datasets). (C-D) The corresponding plots for late-life BDs data from ADNI, RENJI, HUASHAN, and OASIS datasets. Detailed brain patterns are offered in Figs. S3-S4. The detailed data distributions behind B and C are offered in Fig. S2B-C.

## The deep-layer presentation of the DL model for BDs suggests a lifespan spectrum

The model has learned to recruit a set of common features as characterizing dimensions to perform the classification between HC and various BDs. Moreover, deep-layer representations are capable to represent various disorders in a common space with meaningful structure. Then, we explore this deep-layer representation and investigate the possibility of an integrated "lifespan BD spectrum" (Fig. 1E).

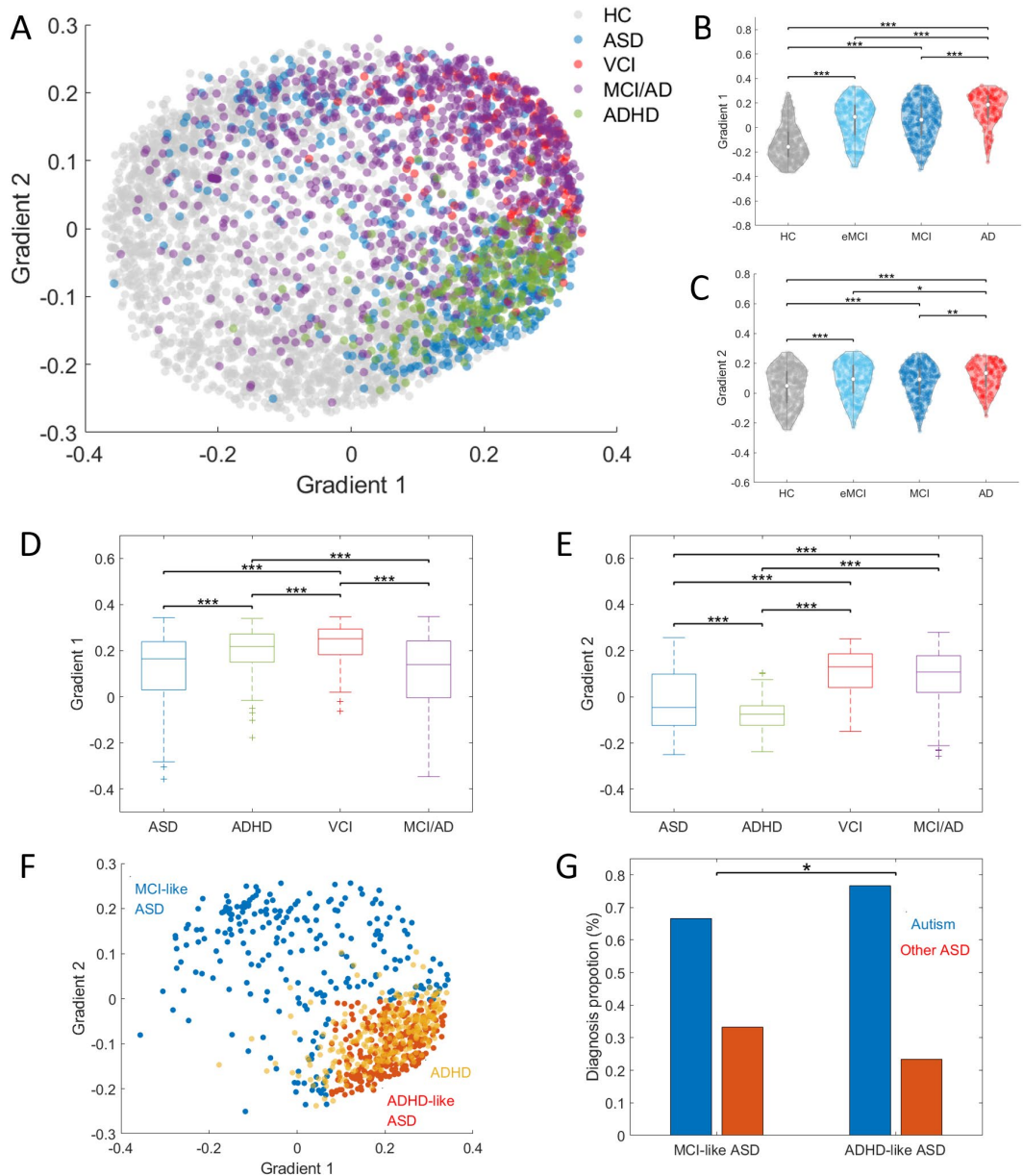

**Figure 5. A low-dimensional representation of the data by the model detects a spectrum for multiple disorders learned.** (A) Two-dimensional space representation for the HC and BD data from the gradient analysis on the deep-layer features from DL models. (B-C) Violin plots for distributions of subjects under AD progression in gradients 1 and 2, respectively. (D-E) Boxplots for ASD, ADHD, VCI, and MCI/AD

distributions in gradients 1 and 2 respectively. + indicates the outliers of the distributions. (F) The locations of ASD and ADHD data in the two-dimensional space. The ADHD-like ASD (being close to ADHD data) and the MCI-like ASD (being close to MCI/AD data) can be identified. (G) Bar plots for diagnosis proportions of autism and non-autism subjects belonging to MCI-like and ADHD-like ASD populations, respectively. In B-G, *: $p<0.05$, **: $p<0.01$, ***: $p<0.001$. Two-sided Whitney-Mann's U test is used in B-E, and the Chi-square test is used in G to generate the $p$-values.

We extract inter-subject relationships as the correlational similarity between individual-level features from the model's deep layer, and decompose these high dimensional relationships using gradient analysis (see Methods). The distribution of the explained variance of the gradients can be found in Fig. S5. The closeness of the BDs in the space, spanned by the first two gradients (called gradient 1 and gradient 2, respectively), largely informs the similarity among BDs in their pathology in terms of RSN alternations (Fig. 5A). It can be first observed that HC and BD subjects are roughly separately distributing in two ends of the two gradients. We use the ADNI data (covering multiple stages of HC-to-dementia progression) to verify these HC-to-BD gradients encoded in our model. In Fig 5B-C, it can be found that, along with the increase of values in the gradients, the diagnosis decisions for subjects gradually change from HC to early-stage MCI (eMCI), late-stage MCI (lMCI), and AD, with statistical significance identified. eMCI and lMCI are not significantly differentiable. These observations support the capability of these gradients to correctly encode the HC-to-BD variation trends. Further in Fig. S6, it can be observed that the variations along gradient 1 and gradient 2 are also associated with age and gender, respectively. Gradient 1 is negatively associated with age with a correlation of $r$=-0.19 and $p$=2.7e-36, while gradient 2 is positively correlated with age with $r$=0.46 and $p$=3.8e-230. And females exhibit a significantly higher value than males in gradient 1, while there is no significant difference in gradient 2.

We then focus on relationships among BDs using closeness in the aligned space (Fig. 5D-E). When ranking BDs based on the population median of the gradient values, gradient 1 depicts a spectrum with the order of MCI/AD, ASD, ADHD, and then VCI, which does not clearly separate early- and late-life BDs. Also, there is not enough evidence to reject the overlapping (i.e., non-significant separation) between ASD and MCI/AD in gradient 1 (two-sided Whitney-Mann's U test, p=1.00, FDR-corrected). Therefore, observations in gradient 1 suggest early- and late-life BDs are connected. Gradient 2 puts ASD and ADHD on one end and MCI/AD and VCI on the other end, which may be due to the strong association between gradient 2 and age (Fig. S6).

Note that there is not enough evidence to reject the overlapping between VCI and MCI/AD in gradient 2 (two-sided Whitney-Mann's U test, p=0.0668, FDR-corrected).

Notably, the ASD data concentrates on two centers in the space, and overlaps *not only* the ADHD data *but also* the MCI/dementia data (Fig. 5F). To explore potentially different traits of these ASD subjects, we first define the ASD data falling within the major distribution of ADHD data (i.e., the 10%-100% percentile in gradient 1, and 0%-90% percentile in gradient 2) as "ADHD-like ASD", and otherwise as "MCI-like ASD". The defined ADHD-like ASDs strongly overlap with ADHD data in gradient 1 (two-sided Whitney-Mann's U test, *p*=0.9848). Based on the diagnosis results from ABIDE datasets, the proportions of autism and other ASD (Asperger's syndrome or pervasive developmental disorder not specified) within ADHD-like ASD and MCI-like ASD are computed and compared (Fig. 5G). The ADHD-like ASD population shows a significantly higher proportion of autism as a diagnosis than the MCI-like ASD population (Chi-square test, *p*=0.0208). This indicates that the ADHD-like ASD population has a higher probability to exhibit autism as the key symptom, while the MCI-like ASD population tends to show other symptoms in the ASD.

## Discussion

In this work, we build a deep learning model, based on a multiscale brain functional network, using 4,410 functional magnetic resonance data, to classify healthy populations from ASD, ADHD, MCI, AD, and VCI with 62.6% accuracy. Previous studies consider identifying different BDs (or different facets of cognitive abilities) using multiple-head architecture under the multi-task framework [35,36]. On the other hand, our work using a single-head architecture is different from those previous studies and provides for the first time a unified viewpoint to investigate BDs from different etiology. However, we admit that the model predictions suffer from inter-individual variations in large data sets, such as ADNI, ABIDE, and ADHD. The model could bias towards late-life BDs than early-life BDs. In addition, due to the limitations in data collection protocol, the VCI group could potentially contain subjects with mixed etiology of AD and vascular diseases.

Despite the drawback due to limited predictability, our model still identifies a set of common features in the BFN, such as connectivity abnormalities with DMN, LIM, ECN, and VIS at different spatial scales, which is consistent with previous findings in several independent studies. For example, ASD, ADHD, MCI, AD, and VCI are all found to be associated with abnormal connectivity in the DMN and ECN [22,37]. And, MCI, AD, and VCI are related to damages within LIM [9,11,23]. This consistency supports the effectiveness of our model learning. According to the neuroscience evidence, the DMN and ECN are related to executive ability (sustained attention and working memory) [22,38], the VIS is associated with the processing of visual information, and the LIM is related to memory storage and retrieval [39–41]. Our RSN-level finding also explains the overlapped behavioral symptoms among BDs, as ASD, ADHD, MCI, AD, and VCI could all exhibit executive ability, visuospatial ability, attention, and memory alterations [18]. These observed neurological factors may also suggest a potential common target for drug delivery and other ways of interferences and treatments in future studies.

In addition, our model learns to represent multiple early- and late-life BDs within a unified space, and the gradient analysis among the data indicates a "lifespan" spectrum connecting disorders with different etiologies. The overlapping among ASD, VCI, and MCI/AD could be in line with observations that VCI and ASD patients have a higher likelihood to develop into AD-type dementia [13]. Also, we observe certain sub-populations of ASD exhibiting similarities to both ADHD and MCI/AD. The analysis of ADHD-like ASD and MCI-like ASD finds the difference in the frequency of occurred symptoms (diagnosis) between the two subpopulations. This could suggest that the lifespan spectrum is a new and informative perspective to review disorders in ASD. Also, it may explain the comorbidities of ASD, ADHD, and MCI/AD [6,19]. In addition, we observed the MCI/AD occupy a large space without obvious concentration in the space. Such variation suggests heterogeneity in BFN deficits underlying the MCI/AD, which is consistent with clinical observations that these brain disorders also show remarkable variations in cognitive manifestations [9,42]. However, note that the currently presented lifespan spectrum could be incomplete as our analysis did not sufficiently include BDs from middle-aged subjects.

The ability of our study in detecting the common pathology in brain dynamics also paves the way for building a general brain-disease diagnosis model. As demonstrated in our results, a model transfer is feasible among BDs. In the field of natural language and image processing filed, pre-training a deep learning model based on multiple source tasks has become a widely-accepted and powerful framework to build generalizable models for multiple downstream tasks, with or without fine-tuning of model parameters [43–45]. However, this pre-training framework has not been widely adopted in the medical image analysis field, and a generalizable pre-trained model is still lacking to fit clinical usages. We hope our study can facilitate the exploration in this direction, towards the development of generalizable artificial intelligence tools for medical imaging applications.

## Methods

### Datasets and Tasks

From four public datasets and two private datasets, we include 4,410 data for training the model, which contains 2512 HC and 1898 BD subjects. The corresponding demographic information is provided in Table S1.

#### Early-life brain disorders

The Autism Brain Imaging Data Exchange (ABIDE) and ADHD-200 datasets, containing neuroimaging from ASD and ADHD subjects, are used for early-life BD identifications.

##### *ABIDE*

From the Autism Brain Imaging Data Exchange (ABIDE-I) [46], we select scans with a duration longer than 300s, yielding 512 HC and 499 ASD subjects. As a multi-site dataset, the acquisition protocols and diagnostic criteria in ABIDE are varying according to data collection sites (16 scan protocols among the sites). Overall, the fMRI scanning parameters are: TR=1.5-3s, TE=15-33ms, in-plane resolution $3\times3$-$3.438\times3.438mm^2$, slice thickness 3-4.5mm, 28-40 axial slices, and 304-486s in duration (120-300 volumes). A detailed protocol can be found at https://fcon_1000.projects.nitrc.org/indi/abide/.

##### *ADHD-200*

From ADHD-200 [47], we use fMRI scans from 488 HC and 280 ADHD subjects, sampled from 8 sampling sites. Again the acquisition protocols significantly vary (9 scan protocols): TR=1.5-2.5s, TE=15-40ms, in-plane resolution $3\times3$-$3.8\times3.8mm^2$, slice thickness 3.0-4.0mm, and 29-47 axial slices. The specific scanning parameters can be found at http://fcon_1000.projects.nitrc.org/indi/adhd200/.

#### Late-life brain disorders

The Alzheimer's disease neuroimaging initiative (ADNI), Open Access Series of Imaging Studies (OASIS), and an in-house HUASHAN dataset contain neuroimaging data from MCI (referring to the prodromal state of AD) or AD elderly subjects. And the in-house RENJI dataset contains data from subjects with VCI. The four datasets are used for investigating late-life BDs.

##### *ADNI*

For the Alzheimer's disease neuroimaging initiative (ADNI) dataset [48], a total of 1350 fMRI data are selected, which contains 565 HC and 785 MCI or AD subjects. Each fMRI data is acquired with TR=3s, TE=30ms, resolution=3.3×3.3×3.3mm$^3$, 48 axial slices, and 420s in duration (140 volumes). A detailed protocol can be found at http://adni.loni.usc.edu/.

*OASIS*

In the Open Access Series of Imaging Studies (OASIS) dataset [49], we include 634 HC and 83 MCI or AD subjects in the study. The fMRI acquisition protocols are TR=2.2s, TE=27ms, resolution=4×4×4mm$^3$, 36 axial slices, and 372s in duration (169 volumes). More information can be found at https://www.oasis-brains.org/.

*HUASHAN*

fMRI data from 167 HC and 100 MCI or AD subjects were obtained from Huashan Hospital in Shanghai[50]. fMRI scans are obtained by using a multi-slice single-shot gradient echo-planar imaging sequence: TR = 0.8s, TE = 37ms, resolution = 2×2×2mm$^3$, 72 axial slices, and 390.4s in duration (488 volumes). The participants are instructed to close their eyes but remain awake during the scanning.

*RENJI*

fMRI data from 146 HC and 151 VCI subjects were obtained from Renji Hospital in Shanghai. MRI scan is performed using a SignaHDxt 3T MRI scanner (GE Healthcare, United States), with an eight-channel standard head coil with foam paddings to restrict head motions. The parameters of the echo-planar imaging sequence for the resting-state fMRI data collection are as follows: TR = 2s, TE = 24 ms, resolution = 2×2×2mm$^3$, 34 axial slices, and 440s in duration (220 volumes). The diagnostic criteria are reported in our previous publications [34,51]. Note that the VCI subjects did not receive positron emission tomography (PET) scans to exclude the AD-related pathology, and thus the VCI group could potentially contain subjects with mixed etiology of both vascular disease and AD.

**fMRI Preprocessing**

We apply well-accepted toolboxes, AFNI [52] (for ADNI) and DPARSF [53] (for ABIDE, OASIS, HUASHAN, and RENJI datasets), to perform a standardized preprocessing procedure for fMRI data. In particular, the first 5 or 10 volumes of each image are discarded due to potential non-equilibrium magnetization. The slice timing correction is done except for the HUASHAN dataset as the data was sampled with high temporal resolution. The rigid-body transformation is performed to correct subjects head motion. Subjects with large head motions are excluded. We do not further perform scrubbing/censoring of data as it may introduce additional artifacts. The signals of white matter, cerebrospinal fluid, and head motion are regarded as nuisance covariates, and are regressed out from individual data. The fMRI images are then normalized to the Montreal Neurological Institute (MNI) space and spatially smoothed with a Gaussian kernel with full width at half maximum (FWHM) of 4×4×4mm$^3$. The BOLD signals are further band-pass filtered (0.01 ≤ f ≤ 0.1 Hz) to remove the neural-irrelevant high-frequency noises and low-frequency drift from MRI machine. For ABIDE dataset, since volumes of scans are different among the collecting sites, we use its minimum common length, i.e., 115 volumes, around the middle volume of the preprocessed fMRI sequence for further processing. We use preprocessed data for ADHD-200 provided in http://preprocessed-connectomes-project.org/adhd200/, using the Athena pipeline.

**Multiscale functional network construction**

Schaefer *et al.* provided a set of atlases for multiscale brain parcellation [31], which are used in this paper for generating multiscale BFNs and guiding the node pooling across scales. The atlases are generated by FC-pattern-based clustering on voxel (or vertices) by considering both global similarity and spatial proximity. Clustering in different resolutions results in brain functional parcellations at multiple scales, ranging from 100 to 1000 regions of interest (ROIs). It can be observed that the seven RSN structures [32] are largely preserved after parcellation at all scales (Fig. 1A). Therefore, the atlases at different scales can be viewed as coarse-to-fine parcellation of the seven RSNs. The spatial relationship among the ROIs in these atlases at different scales thus characterizes a biologically-meaningful functional hierarchy.

Given the atlas at a specific scale, the ROI-level signals can be obtained by averaging voxel-level BOLD signals within each ROI. The BFN at the given scale $S$ is then computed by Pearson correlation among all pairs of ROI-level signals, and is denoted as $BFN_S$. Consistent with our previous study [30], we use the first five scales, i.e., from 100 to 500 ROIs.

**Multi-site data harmonization**

To deal with data inconsistency among multiple datasets, a statistical regression-based harmonization method, called "Combat" [54,55], is applied to calibrate the BFN data. The codes are publicly available at https://github.com/Jfortin1/ComBatHarmonization. In Combat, with a linear regression model, the variation of each functional connectivity across individuals is modeled as the sum of essential mean, effects of biological co-variates (i.e., age, gender, and brain BDs), site-related bias in mean, and site-related noise level. Therefore, functional connectivity without site effect can be calculated by estimating the parameters of the regression model from data and removing site-related bias.

In this paper, we use a scanner-based harmonization since ABIDE and ADHD-200 contain data from multiple sites (ABIDE: 16 scanners; ADHD: 9 scanners; In total, 28 scan protocols for all data). We preserve the effect of age, gender, and type of BDs.

**Deep learning architecture**

The Multiscale-Atlas-based Hierarchical Graph Convolutional Neural Network (MAHGCN) is proposed and systematically tested in our previous study [30]. Here we briefly review two crucial building blocks of MAHGCN, i.e., graph convolutional network (GCN) and the atlas-guided pooling (AP). The MAHGCN is then built by hierarchically stacking GCNs and APs (Fig. 1C), together with the skip connections and fully-connected layers (FLs).

Graph Convolutional Network

The graph convolutional network (GCN) [56] is an effective deep-learning method to abstract features from graph data (e.g., the BFN data). It completes the convolutional operations via two steps, i) propagating nodal features via graph Laplacian, and ii) selecting features by applying a learned kernel on the features. Formally, for a given adjacency matrix $A$ and nodal features $h$, one graph convolution layer updates the nodal feature by following the equation below:

$$h^{GC} = \text{C} \ (A, h) = \sigma\left(\widetilde{D}^{-\frac{1}{2}}\widetilde{A}\widetilde{D}^{-\frac{1}{2}}hW\right),\tag{1}$$

where $\widetilde{A} = A + I$, $I$ is the identity matrix, $\widetilde{D}$ is the corresponding degree matrix of $\widetilde{A}$, $W$ is the estimated kernel weight matrix, and $\sigma(\cdot)$ is a non-linear activation function. Empirically, we skip the computation of graph Laplacian and directly use the adjacency matrix $BFN_S$ from scale $S$ to obtain optimal diagnosis performance.

Atlas-guided Pooling

The AP operation is defined according to spatial overlapping among ROIs informed by atlases at different scales. The AP benefits information integration and introduces inter-scale dependency during feature extraction. It aims to convert the nodal features defined by the atlas at scale $P$ into the nodal features for the atlas at scale $Q$ ($P > Q$), based on the mapping matrix $M_{P \rightarrow Q}$:

$$M_{\mathcal{R} \rightarrow \wp}(i,j) = \begin{cases} 1, & \rho > Th \\ 0, & \text{Otherwise} \end{cases},\tag{2}$$

where the overlapping ratio $\rho$ is computed by size (i.e., the number of voxels) of spatially overlapping between ROI $i$ in the atlas at scale $P$ and ROI $j$ in the atlas at scale $Q$ divided by the size of ROI $i$. And $Th$ is a threshold applied to $\rho$ for defining elements in $M_{P \rightarrow Q}$. We use $Th = 0$ according to the results in our previous methodological paper [30]. Through a matrix multiplication with $M_{P \rightarrow Q}$, a feature map $h_P^{GC}$ defined in the atlas at scale $P$ from GCN is converted into a new feature map $h_Q^{AP}$ for the atlas at scale $Q$.

<u>Implementation</u>

All models are implemented using the open-source framework "Pytorch" in Python. During implementation, we choose the ReLU function as the non-linear activation function. An identity matrix is used as the initial nodal feature to make the MAHGCN model focus on the topology of the BFN. The GCN layers in MAHGCN are attached with dropout functions (rate = 0.3), and the last GCN layer is followed by four FLs. Each FL is associated with a batch normalization and a ReLU activation function. The outputs from the last (the 4th) FL are normalized by a Softmax function to generate the diagnostic probabilities for two classes. These configurations for GCN and FL are kept consistent in the single-scale-based GCN methods.

**Bi-classification experiments**

First, the MAHGCN model is used to classify all BDs from HCs. As described above, the model is restricted to be a single-head architecture and thus forced to extract one set of features being diagnostic for all BDs. Thus, a successful classification demonstrates common features among all BDs.

<u>Training scheme</u>

Since sample size and class ratio (i.e., HC-vs-BD ratio) are different in each dataset, a site-specific weight and a cross-entropy loss function are used to supervise the training process. The weighted cross-entropy loss is based on the inverse of the HC-vs-disorder ratio for each site, estimated in the training samples. For each update iteration, we randomly sample (equally, 100 samples) from each set, which are inputted to the model to calculate their site-specific losses, respectively. The yielded site-specific cross-entropy loss is further multiplied with a penalty designed by the square root of the inverse of the site sample size for re-weighting. All re-weighted site-specific losses are accumulated with the linear summation, based on which the model parameters are finally updated.

The training parameters for neural network models are identically set as training epoch

= 150, and learning rate = 0.01 for the first 50 epochs and then 0.001 for the remaining epochs. Adam [57] with a weight decay of 0.01 is used as an optimizer. Other parameters of the neural network models are initialized with random weights with the default setting of Pytorch.

<u>Validation scheme</u>

A classic ten-fold cross-validation is performed. The data is randomly shuffled and equally split into ten folds. In each round of cross-validation, nine folds of data will be used as training samples and the remaining one as testing samples. Ten rounds of cross-validation are performed until all folds play as testing samples once. Four metrics are adopted to evaluate performance in the testing samples, i.e., accuracy (ACC), sensitivity (SEN), specificity (SPE), and area under the receiver operating characteristic curve (AUC). Since sample sizes in different disorders from different sites are significantly varying, computing "global" statistics simply as the ratio of correct predictions against all samples will assign larger weights to the sites with larger sample sizes. We thus compute four performance metrics for each site ("site-specific" statistics) and then average over all sites ("site-averaged" statistics) for each cross-validation. The mean and standard deviation of "site-specific" statistics and "site-averaged" statistics from cross-validation are reported.

**Transfer learning experiments**

We tested whether a model pre-trained using all data except ABIDE dataset ($N$=3399) can be transferred to perform ASD identification in ABIDE data ($N$=1011) with restricted samples. This experiment aims to provide additional evidence for the common features under different BDs.

<u>Training scheme</u>

The model is trained on the five datasets until it converges with 250 epochs. This pre-trained model is used as an initial model and further fine-tuned using training samples from ABIDE with 50 epochs. For both pre-training and fine-tuning, other configurations are the same as the settings in bi-classification experiments. In addition, four levels of fine-tuning schemes are designed to test the model with different amounts of preservation of the learned information during the pre-training. "Level 1" refers to fine-tuning all model parameters. "Level 2" refers to fine-tuning all FLs and batch normalization layers (BN). "Level 3" refers to fine-tuning the last FL and all the BNs in the model. "Level 4" refers to fine-tuning only the last FL and the last BN. Intuitively, higher level fine-tuning preserves more learned information during the pre-training.

<u>Validation scheme</u>

A ten-fold "K-shot" cross-validation is performed. For each round of cross-validation, the data are shuffled and split into training ($N$=100) and testing sets ($N$=911). For the K-shot condition, training samples are the first K samples in the training set. In the main text, results using a 20-shot condition are depicted. In Tables S2-3, we offer the results under 50 shots and 100 shots, which are consistent with the results under the 20-shot condition. The mean and standard deviation of ACC, SEN, SPE, and AUC in the testing set is used to assess the performance.

**Diagnostic feature identification**

To reveal the predictive features of deep learning methods, we utilize a Gradient-guided Class Activation Map (Grad-CAM) algorithm [58] and analyze the established bi-classification models. In short, the Grad-CAM regards the gradient between prediction outputs and the feature maps at intermediate hidden layers (in this work, we used features from intermedia GCN layers for each scale) of the deep neural network as the importance of features. This thus applies gradient values to weight elements in the feature maps, i.e., the product between the gradient map and feature map, namely the class activation map (CAM), which offers a visual map for spotting predictive features.

In order to investigate the common features of BDs, the Grad-CAM from correctly predicted BD subjects is extracted using different models from cross-validations. However, the Grad-CAM values can vary significantly across different models and different datasets due to the nature of using multi-sites. Therefore, we designed a double normalization procedure to relieve the Grad-CAM value heterogeneity across the models and datasets to spot the common features more properly. First, all Grad-CAM values from a given model are normalized into a range from zero to one according to the minimum and maximum values. Then, all normalized Grad-CAM from subjects belonging to different datasets is averaged respectively. To obtain a joint estimation of the models from cross-validations, we utilize the prediction AUCs to perform a weighted average on the normalized Grad-CAM. In this way, the normalized Grad-CAMs for every specific dataset are established. We again normalize these dataset-specific Grad-CAMs into a range from zero to one to address amplitude differences in Grad-CAMs across datasets. For the all-BD common features (Fig. 3), we average all six double-normalized Grad-CAMs. For common features of early-life BDs (Fig. 4A), the double-normalized Grad-CAMs from ABIDE and ADHD-200 are averaged. And for common features of late-life BDs (Fig. 4B), the double-normalized Grad-CAMs from ADNI, OASIS and HUASHAN, and RENJI are averaged.

**Estimation of the spectrum representation under various BDs**

To explore the existence of a common spectrum under different BDs, we investigate deep-layer data representations from the established bi-classification models. The encoded features of individuals are extracted from third-layer FL, and the sample relationships are computed by the correlation distance based on these features. The sample relationship from models under different rounds of validation is then weighted-averaged using AUCs to provide a common estimation, based on which individuals are embedded into a low-dimensional Euclidean space with principal coordinate analysis. Specifically, we used the diffusion map method, which is in alignment with the brain gradient analysis [59]. The diffusion map method estimates a non-linear mapping of the data into a new low-dimensional Euclidean space to ensure a distance-preserved mapping, so that the Euclidean distances among individuals in the mapped space roughly keep the original distances reflected in the sample relationship matrix. The implementation of the diffusion map is based on the open source "BrainSpace" toolbox (http://github.com/MICA-MNI/BrainSpace) [60], with default settings.

**Statistical analysis**

The differences in performance from different methods are tested by the Wilcoxon signed-rank test using a build-in function "signrank" in Matlab (Tables 1 and 3). Other comparisons are performed by Whitney-Mann's U test with "ranksum" in Matlab. Both Wilcoxon signed-rank test and Whitney-Mann's U test are non-parametric. Under multiple comparisons, the raw $p$-values are corrected by the false discovery rate (FDR) correction.

To assess the significance of predictability during the bi-classification, we conduct permutation to randomize the ground-truth labels, and re-calculate the performance metrics to estimate the corresponding distribution under chance level. As we use the ten-fold cross-validation, 100 times permutations are conducted using the results from one round of cross-validation. The results from the 1000 permutations are then pooled to generate the estimation of the chance-level distribution. The significance ($p$-value) is then obtained by statistically comparing the empirical distribution from trained models and the chance-level distribution, using a one-sided Whitney-Mann's U test.

## Acknowledgments

This work was supported in part by the National Natural Science Foundation of China (grant number 62131015), the Science and Technology Commission of Shanghai Municipality (STCSM) (grant number 21010502600), Shanghai Leading Talent Program (grant number 2022LJ023), National Natural Science Foundation of China (grant number 82171885), and Shanghai Science and Technology Committee Project (Natural Science Funding, grant number 20ZR1433200; Explorer Program, grant number 21TS1400700).

## Author Contributions

M.L., Q.W., H.Z., and D.S designed the research; Y.W., Y.Z., F.X., and Q.G. collected the data; M.L and J.Z. wrote the code, analyzed data, and drafted the initial manuscript; all authors revised the paper.

## Competing Interests

The authors declare no competing interests.

## Materials and Correspondence

Correspondence and material requests are addressed by D.S.

## Data availability

Data from ADNI, OASIS, ABIDE, and ADHD-200 datasets are publicly available. Data from RENJI and HUASHAN datasets are available from the corresponding authors upon reasonable request due to privacy restrictions.

## Code availability

Codes are available from the corresponding authors upon request during the review process and will be made publicly available upon acceptance.

Supplementary Materials for

**Deep learning reveals the common spectrum underlying multiple brain disorders in youth and elders from brain functional networks**

Mianxin Liu, Jingyang Zhang, Yao Wang, Yan Zhou, Fang Xie, Qihao Guo, Feng Shi, Han Zhang, Qian Wang and Dinggang Shen*

*Corresponding author. Email: Dinggang.Shen@gmail.com

**This PDF file includes:**

Tables S1 to S3
Figures. S1 to S6

Table S1. Demographic information for the subjects in each dataset.

| | Gender (M/F) | | Age (Years) | |
|---|---|---|---|---|
| | HC | Disorder | HC | Disorder |
| ABIDE | 434/78 | 436/63 | 17.6±7.7 | 17.2±8.5 |
| RENJI | 118/28 | 110/41 | 65.5±7.2 | 65.1±7.2 |
| PET CENTER | 62/105 | 41/59 | 63.9±7.9 | 64.9±7.0 |
| ADNI | 235/330 | 418/367 | 74.6±7.3 | 73.5±7.6 |
| OASIS | 276/358 | 46/37 | 64.4±8.7 | 71.8±7.2 |
| ADHD | 258/230 | 220/60 | 12.2±3.3 | 11.6±3.0 |

Table S2. The prediction performances in transfer learning experiments under baseline (no pre-training) and different transfer learning schemes (pre-trained). The results are from the 50-shot condition. * indicates the performance metrics are significantly higher than baseline at a significance level of $p<0.05$. **: $p<0.01$ and ***: $p<0.001$ after FDR correction. A one-sided Wilcoxon signed-rank test is used. Related to Table 3.

| Scheme | ACC | $p$ | SEN | $p$ | SPE | $p$ | AUC | $p$ |
|---|---|---|---|---|---|---|---|---|
| Baseline | 53.10±1.2 | N/A | 45.84±7.5 | N/A | 60.24±8.2 | N/A | 53.04±1.2 | N/A |
| Level-1 | 52.25±1.1 | 1.0 | **66.38±9.7**\*\* | 0.0039 | 38.35±10.8 | 1.0 | 52.37±1.1 | 1.0 |
| Level-2 | 52.76±0.9 | 0.9427 | 58.39±8.2** | 0.0039 | 47.22±9.3 | 1.0 | 52.80±0.8 | 1.0 |
| Level-3 | 54.50±0.8* | 0.0391 | 47.61±5.2 | 0.6523 | **61.28±3.7** | 0.3281 | 54.45±0.8* | 0.0273 |
| Level-4 | **57.14±0.2**\*\* | 0.0039 | 56.33±3.1** | 0.0020 | 57.93±3.2 | 0.5771 | **57.13±0.2**\*\* | 0.0013 |

Table S3. The prediction performances in transfer learning experiments under baseline (no pre-training) and different transfer learning schemes (pre-trained). The results are from the 100-shot condition. * indicates the performance metrics are significantly higher than baseline at a significance level of $p<0.05$. **: $p<0.01$ and ***: $p<0.001$ after FDR correction. A one-sided Wilcoxon signed-rank test is used. Related to Table 3.

| Scheme | ACC | $p$ | SEN | $p$ | SPE | $p$ | AUC | $p$ |
|---|---|---|---|---|---|---|---|---|
| Baseline | 54.17±0.9 | N/A | 42.34±11.1 | N/A | **65.81**±11.3 | N/A | 54.08±0.9 | N/A |
| Level-1 | 53.30±0.6 | 0.9912 | 51.84±4.6 | 0.2617 | 54.75±4.9 | 1.0 | 53.29±0.5 | 1.0 |
| Level-2 | 51.91±0.5 | 1.0 | 47.28±6.3 | 1.0 | 56.47±6.5 | 1.0 | 51.87±0.4 | 1.0 |
| Level-3 | 55.66±0.5** | 0.0039 | 57.35±1.4** | 0.0039 | 54.01±1.4 | 0.9941 | 54.45±0.8** | 0.0039 |
| Level-4 | **56.53**±0.2** | 0.0039 | **59.96**±1.3** | 0.0026 | 53.16±1.3 | 0.9971 | **56.56**±0.2** | 0.0020 |

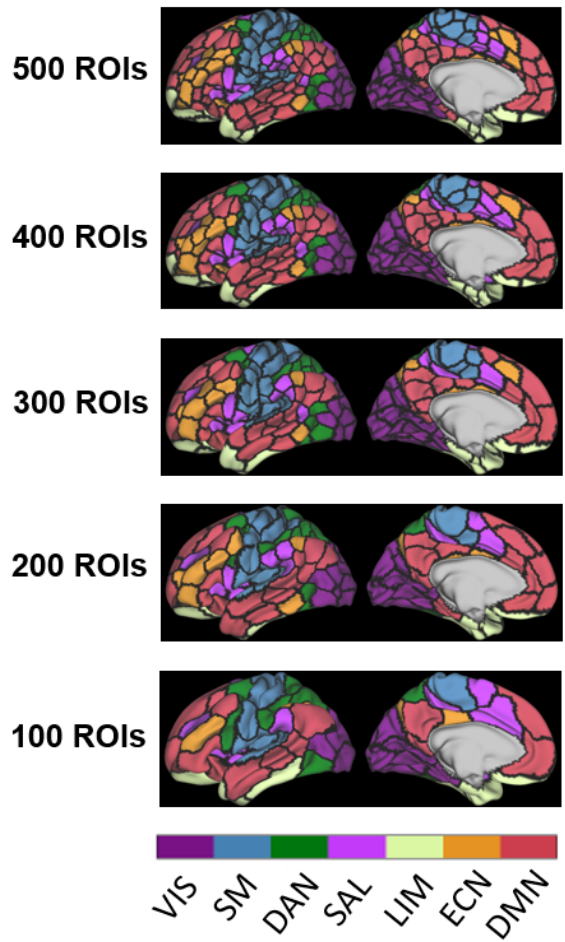

Figure S1. The implemented multiscale atlases from Schaefer et al. The colors on the surface of the atlases indicate the belongings of the regions to different resting-state networks (RSNs), including visual network (VIS), somatomotor network (SM), dorsal attention network (DAN), salience network (SAL), limbic network (LIM), executive control network (ECN) and default mode network (DMN).

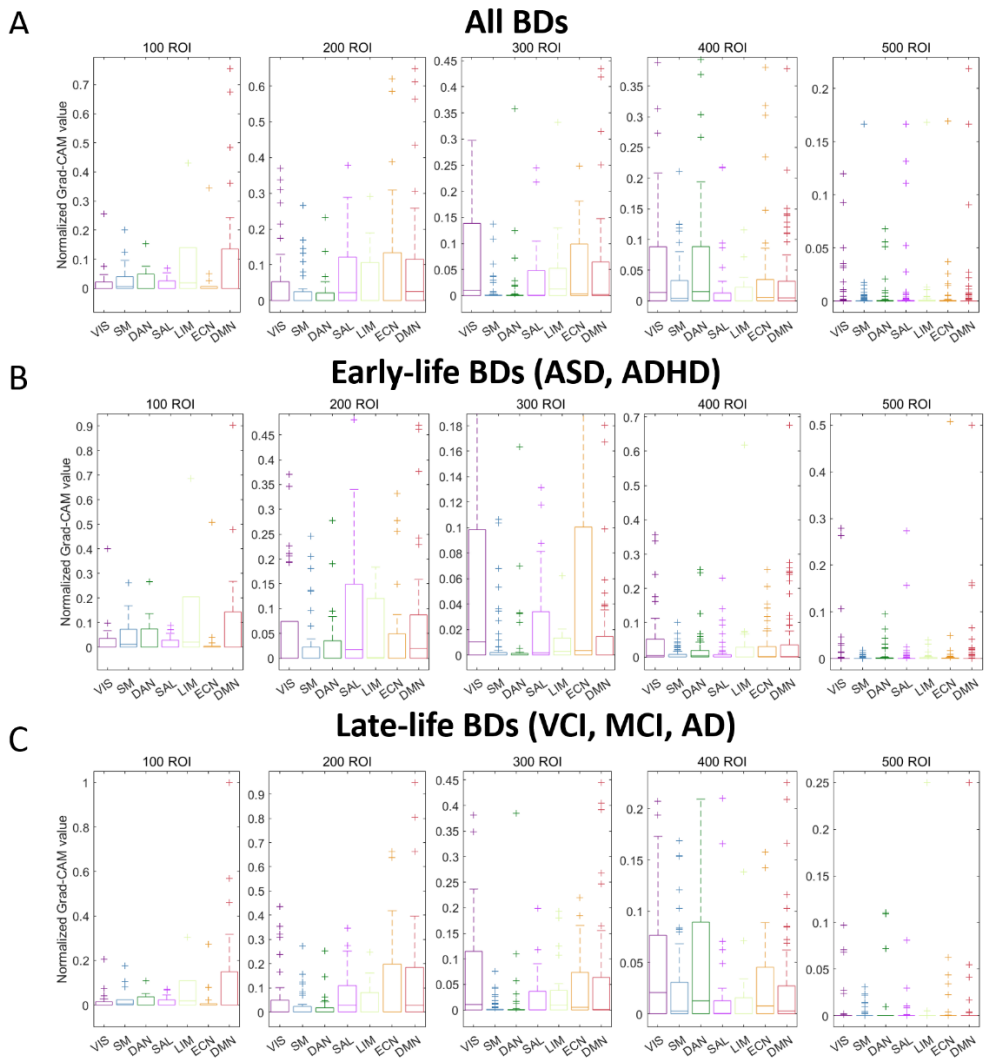

Figure S2. Detailed distributions of diagnostic features in the resting state networks. (A). Boxplot for results overall brain disorders, also as the data behind Fig. 3B. (B). For early-life brain disorders also as the data behind Fig. 4B. (C). For late-life brain disorders, also as the data behind Fig. 3D. Related to Figs. 3 and 4.

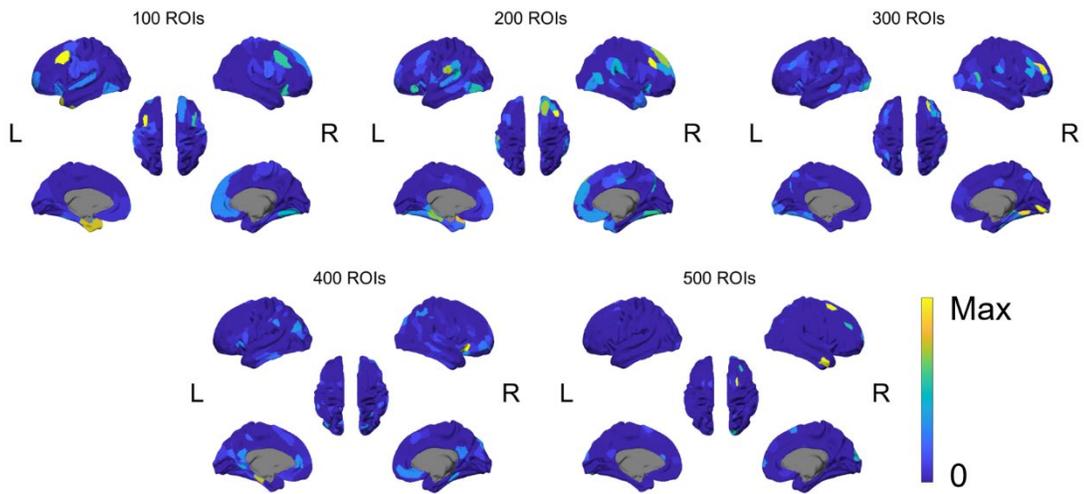

Figure S3. Detailed distributions of diagnostic regional features for early-life brain disorders as brain maps, from ABIDE and ADHD-200 datasets. Related to Fig. 4.

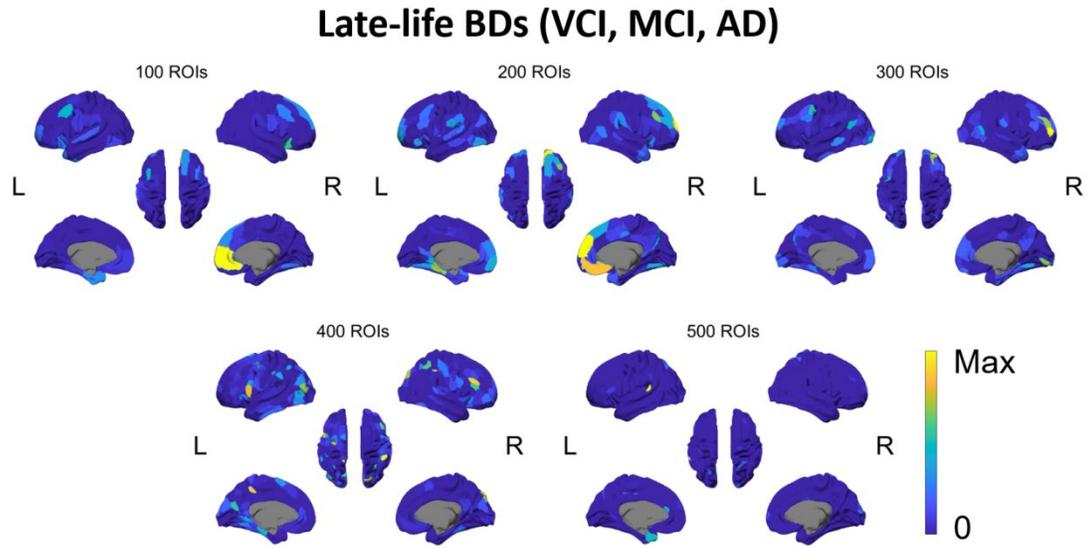

Figure S4. Detailed distributions of diagnostic regional features for late-life brain disorders as brain maps, from ADNI, OASIS, RENJI, and HUASHAN datasets. Related to Fig. 4.

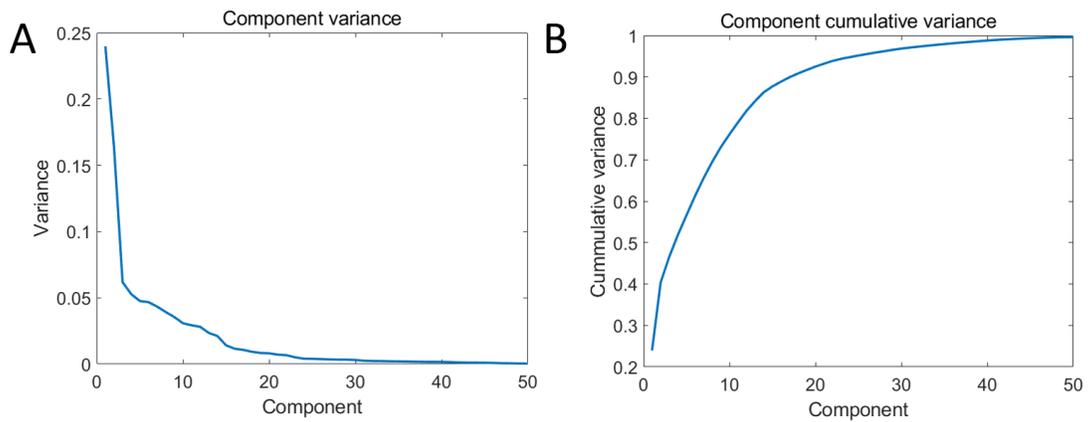

Figure S5. The variance (λ) of the diffusion embedding components. (A). The variance for each component. (B). The accumulative variance for each component. Results from the first 50 components are depicted. Related to Fig. 5.

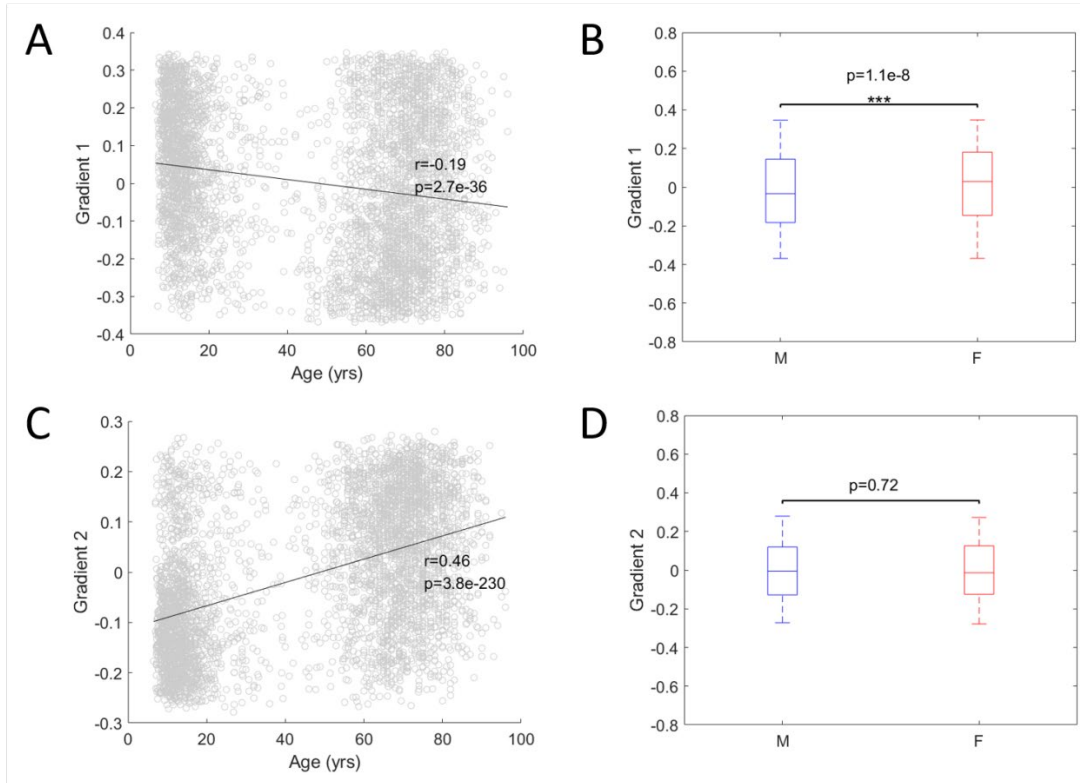

Figure S6. Dependencies of age, gender, and variation in gradients 1 and 2. Related to Fig. 5.